\documentclass[journal]{IEEEtran}
\usepackage{amsmath,amsfonts,amssymb}
\usepackage{algorithmic}
\usepackage{algorithm}
\usepackage{array}
\usepackage[caption=false,font=normalsize,labelfont=sf,textfont=sf]{subfig}
\usepackage{textcomp}
\usepackage{stfloats}
\usepackage{url}
\usepackage{verbatim}
\usepackage{graphicx}
\usepackage{cite}

\usepackage{amssymb}
\usepackage{bbding}
\usepackage{xcolor}
\usepackage{colortbl}

\usepackage[normalem]{ulem}  
\usepackage{xcolor}

\def\ourmodel{CamoSAM2}  
\def\ournet{MAPI}  
\def\ourAlgorithm{AMPR} 

\def\eg{\emph{e.g.}}

\def\ie{\emph{i.e.}}

\definecolor{indiagreen}{rgb}{0.07, 0.53, 0.03}
\definecolor{mygray}{gray}{.92}

\definecolor{revisioncolor}{RGB}{180,30,30}

\usepackage{hyperref}
\hypersetup{
colorlinks=true,
linkcolor=red, 
}
\usepackage{multirow}
\usepackage{booktabs}
\begin{document}

\title{CamoSAM2: SAM2-oriented Prompt Auto-Refinement for Video Camouflaged Object Detection}

\author{Xin~Zhang,~~Keren~Fu,~~and~Qijun~Zhao

\IEEEcompsocitemizethanks{\IEEEcompsocthanksitem 
Manuscript received on August 11, 2025. 
\textit{(Corresponding author: Keren Fu.)}
\IEEEcompsocthanksitem Xin Zhang is with the National Key Laboratory of Fundamental Science on Synthetic Vision, Sichuan University, Chengdu 610065, China. (E-mail: zhangxinchina1314@gmail.com)
\IEEEcompsocthanksitem Keren Fu and Qijun Zhao are
with the College of Computer Science, and
the National Key Laboratory of Fundamental Science on Synthetic Vision,
Sichuan University, Chengdu 610065, China. (E-mail: fkrsuper@scu.edu.cn; qjzhao@scu.edu.cn)\protect
}
}

\markboth{Journal of \LaTeX\ Class Files}%
{Zhang \MakeLowercase{\textit{et al.}}: Bare Demo of IEEEtran.cls for IEEE Journals}

\maketitle

\begin{abstract}
The Segment Anything Model 2 (SAM2), a prompt-guided video foundation model, has remarkably performed in video object segmentation, drawing significant attention in the community.
Due to the high similarity between camouflaged objects and their surroundings, which makes them difficult to distinguish even by the human eye, the application of SAM2 for automated segmentation in real-world scenarios faces challenges in camouflage perception and reliable prompts generation.
To address these issues, we propose \ourmodel, a motion-appearance prompt inducer (\ournet) and refinement framework to automatically generate and refine prompts for SAM2, enabling high-quality automatic detection and segmentation in VCOD task.
Initially, we introduce a prompt inducer that simultaneously integrates motion and appearance cues to detect camouflaged objects, delivering more accurate initial predictions than existing methods.
Subsequently, we propose a video-based adaptive multi-prompts refinement (AMPR) strategy tailored for SAM2, aimed at mitigating prompt error in initial coarse masks and further producing good prompts. Specifically, we introduce a novel three-step process to generate reliable prompts by camouflaged object determination, pivotal prompt frame selection, and multi-prompts formation. 
Extensive experiments conducted on two benchmark datasets demonstrate that our proposed model, \ourmodel, significantly outperforms existing state-of-the-art methods, achieving increases of 8.0\% and 10.1\% in mIoU metric. 
Additionally, our method achieves the fastest inference speed compared to current VCOD models.
The code will be made publicly available at \url{https://github.com/zhangxin06/CamoSAM2}.
\end{abstract}


\section{Introduction}
\IEEEPARstart{C}{AMOUFLAGED} 
object detection (COD) seeks to identify and segment \textit{hidden objects} that blend seamlessly into their surroundings. This task is critical in computer vision with a wide range of applications, including surveillance~\cite{liu2019concealed}, medical image analysis~\cite{fan2020pra,wu2021jcs,ji2022vps,wu2023medical}, and wildlife conservation~\cite{lidbetter2020search}. 
While significant progress has been made in detecting camouflaged objects from a single image, some camouflages in nature remain nearly imperceptible in static scenes, even to the most perceptive predators. However, once the concealed prey moves, the concealment is disrupted, making them susceptible to a predator’s attack.
This natural phenomenon has inspired research into harnessing motion cues to tackle the challenges of camouflaged object detection in videos \cite{yang2021selfsupervised,cheng2022implicit,hui2024implict-explicit,zhang2025emip}. While these methods have shown encouraging results, the limited availability of training data often results in overfitting and poor generalization to unseen scenarios.
Recently, the visual foundation models such as the Segment Anything Model (SAM)~\cite{kirillov2023segment} and its video extension SAM2~\cite{ravi2024sam2} have demonstrated strong generalization capabilities, offering a promising direction for VCOD.

\begin{figure}[t!]
  \centering
  \includegraphics[width=0.48\textwidth]{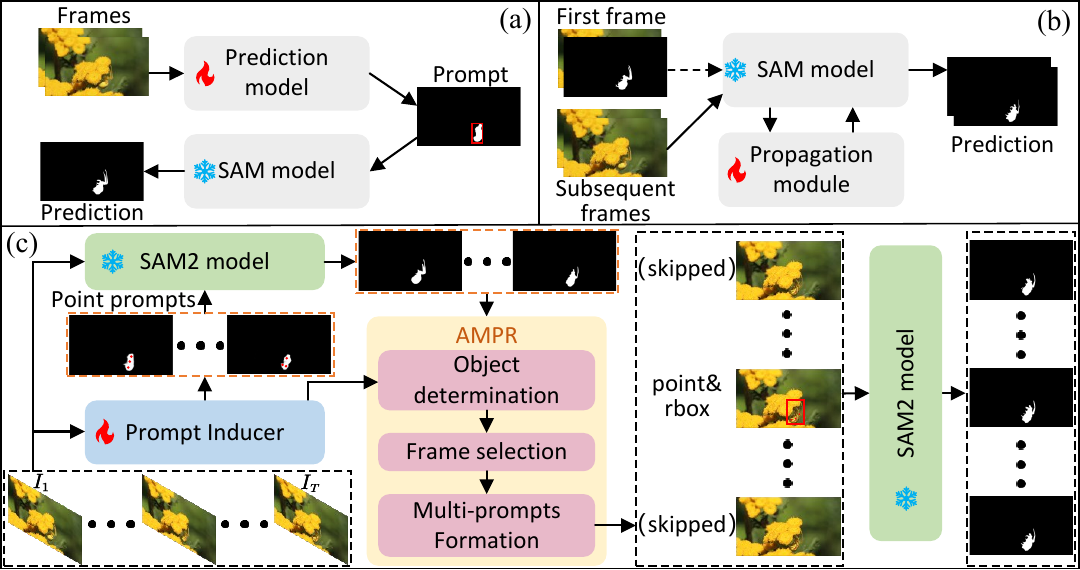}
  \caption{Illustration of previous SAM-based VCOD methods (a-b) compared with ours (c): 
  (a) Generating prompts and directly feeding into SAM \cite{hui2024endow}; (b) Incorporating a newly designed module into SAM to propagate initial user-provided masks across subsequent frames \cite{meeran2024sam}; 
  (c) Our approach, which first derives coarse prompts and then automatically refines them to improve the effectiveness of SAM2.}
  \label{fig: figure1}
\end{figure}

As illustrated in Fig.~\ref{fig: figure1}, existing SAM-based approaches mainly follow two paradigms. The first line of work (Fig.~\ref{fig: figure1}(a)) generates prompts (e.g., masks or bounding boxes) from video frames and feeds them into SAM for segmentation~\cite{hui2024endow}. While this eliminates manual interaction, it treats each frame independently and fails to model temporal consistency, essentially degenerating into frame-wise segmentation. The second line of work (Fig.~\ref{fig: figure1}(b)) leverages SAM as a feature backbone and incorporates propagation mechanisms to maintain temporal coherence~\cite{meeran2024sam}. However, such approaches rely on high-quality user-provided masks in the first frame, which is impractical in fully automatic scenarios.

SAM2 extends SAM to video settings and enables mask propagation across frames. Nevertheless, its performance critically depends on the quality of the input prompts. In VCOD scenarios, generating reliable prompts is particularly challenging due to inaccurate localization, incomplete masks, and severe appearance ambiguity. As a result, directly applying SAM2 without carefully designed prompt strategies often leads to suboptimal performance.
In this work, we argue that \textbf{prompt quality is the central bottleneck in SAM2-based VCOD}, and identify two key challenges. 
\textbf{(1) Localization ambiguity:} existing VCOD models can provide coarse mask predictions as prompts, but these masks are often inaccurate and incomplete due to weak global reasoning, limiting their effectiveness. 
\textbf{(2) Temporal instability:} under challenging conditions such as occlusion, motion blur, and scene transitions, the quality of these prompts becomes unreliable, further degrading segmentation performance.

To address these challenges, we propose a fully automatic SAM2-based framework that focuses on \textit{prompt generation and refinement}. Specifically, we introduce a more accurately positioned motion-appearance prompt inducer, termed MAPI, which integrates motion and appearance cues to improve object localization and generate coarse but informative mask prompts. Importantly, MAPI is designed as an efficient and replaceable module, making it compatible with existing VCOD pipelines.
Furthermore, to enhance robustness under dynamic scenarios, we propose an adaptive multi-prompts refinement module, AMPR. This module operates in a training-free manner by selecting pivotal frames based on temporal consistency and refining prompts using complementary box and point cues. By filtering unreliable predictions and reinforcing high-quality prompts, AMPR effectively stabilizes the input to SAM2, leading to more reliable segmentation results.

By explicitly addressing both localization quality and temporal robustness, the proposed framework enables SAM2 to perform accurate and consistent segmentation in VCOD without requiring any human interaction.

To the best of our knowledge, this is the first work that systematically studies prompt generation and refinement for SAM2 in VCOD and provides a fully automated solution.

The main contributions are summarized as follows:
\begin{itemize}
    \item We identify prompt quality as the key bottleneck in SAM2-based VCOD and propose a unified framework that explicitly addresses both localization accuracy and temporal robustness.

    \item We propose a more accurately positioned motion-appearance prompt inducer, MAPI, which improves object localization and generates informative mask prompts by integrating motion and appearance cues.

    \item We introduce a training-free adaptive multi-prompts refinement module, AMPR, which enhances prompt reliability by selecting pivotal frames and refining prompts based on temporal consistency.

    \item Extensive experiments demonstrate that our method achieves significant improvements over state-of-the-art approaches, validating the effectiveness of our design.
\end{itemize}

\begin{figure*}[t!]
\begin{center}
\includegraphics[width=1.0\textwidth]{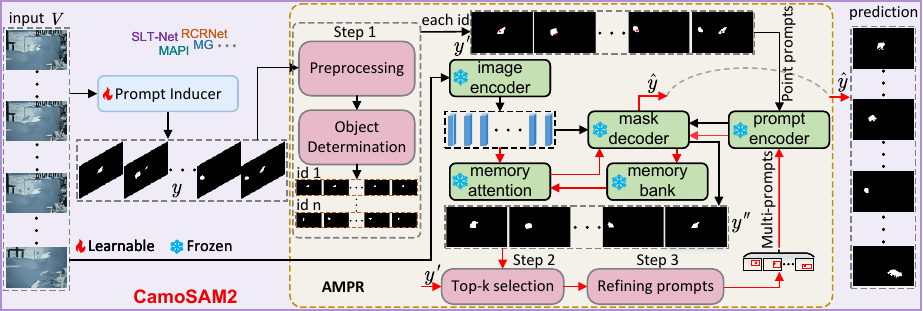}
\end{center}
\vspace{-8pt}
\caption{\small
Pipeline of \ourmodel, which consists of two main components: a replaceable prompt inducer (\eg, SLT-Net~\cite{cheng2022implicit}, RCRNet~\cite{yan2019semi}, MG~\cite{yang2021selfsupervised}, \ournet~(Ours),$\cdots$) and video-based adaptive multi-prompts refinement (AMPR). The fire and snowflake symbol signifies that the model parameters in this part are kept learnable and frozen, respectively.
Specifically, the red arrow represents the data flow from Step 2 to the final predicted result $\hat{y}$.
}
\label{fig:framework}
\end{figure*}

\section{Related Work}
\subsection{Image-based COD} 
Methods in this category focus on detecting camouflaged objects within a single RGB image. Inspired by natural predatory behaviors, approaches such as SINet-V2~\cite{fan2021concealed} and PFNet~\cite{mei2021camouflaged} employ a coarse-to-fine strategy. These methods initially generate a preliminary map to locate potential camouflaged objects, which is then progressively refined for accurate segmentation. 
To further improve detection performance, Zhai~\textit{et al.}~\cite{zhai2021mutual} introduced an auxiliary task that integrates classification or boundary detection with camouflaged object detection. Additionally, Jia~\textit{et al.}~\cite{jia2022segmar} proposed the SegMaR framework, an iterative refinement approach designed to locate, magnify, and detect camouflaged objects. 
Xing~\textit{et al.}~\cite{xing2023SARNet} proposed a three-stage Search-Amplify-Recognize framework, which leverages attention-driven localization, feature amplification, and accurate figure-ground separation to address the task of COD.
Ren~\textit{et al.}~\cite{ren2023TAF} introduced multiple texture-aware refinement modules to capture subtle texture differences between camouflaged objects and the background.
Wu~\textit{et al.}~\cite{wu2025conet} introduced a consistency-oriented network that models visual consistency between foreground and background through primary detection, consistency evaluation, and detail refinement modules.
Khan~\textit{et al.}~\cite{khan2024camofocus} introduced a feature split and context refinement network to refine camouflaged features.
Yao~\textit{et al.}~\cite{Yao2024HGINet} designed a graph interaction network to discover camouflaged objects effectively.
Several studies~\cite{he2023FEDER,cong2023frequency,zhong2022detecting,sun2024frequency,Liu2024frequency} leveraged frequency-learning modules to extract subtle foreground-background cues by decomposing features into multiple frequency components, thus enhancing spatial identification.
Different from image-based COD studies, our work targets the more practical yet challenging video setting, where temporal motion provides crucial cues to break camouflage.

\subsection{Video-based COD}
For the VCOD task, motion cues are essential for effective detection.
Bidau~\textit{et al.}~\cite{bideau2016s} introduced a method by approximating various motion models derived from dense optical flow. 
Zhang~\textit{et al.}~\cite{zhang2017bayesian} proposed a camouflage modeling (CM) strategy by jointly modeling foreground and background representations, and fusing CM with discriminative (DM) in a Bayesian framework for camouflaged object detection.
Lamdouar~\textit{et al.}~\cite{lamdouar2020betrayed} introduced a video registration and segmentation network for detecting camouflaged objects, leveraging optical flow and difference images as inputs. However, the reliance on imprecise optical flow can lead to cumulative errors in mask prediction.
To address this issue, Cheng~\textit{et al.}~\cite{cheng2022implicit} developed a two-stage model that implicitly captures and utilizes motion information. 
Subsequently, to eliminate inaccuracies stemming from implicit motion modeling in SLT-Net~\cite{cheng2022implicit}, Hui~\textit{et al.}~\cite{hui2024implict-explicit} introduced a motion-induced consistency preserving approach between frames with a feature pyramid framework.
Zhang~\textit{et al.}~\cite{zhang2025emip} introduced a novel explicit motion handling and interactive prompt framework named EMIP to simultaneously facilitate motion estimation and object segmentation.
More recently, efforts have been directed toward adapting SAM for VCOD tasks. Hui~\textit{et al.}~\cite{hui2024endow} leveraged temporal and spatial relationships between frames to generate mask and bounding box prompts for interaction with SAM. 
Additionally, Meeran~\textit{et al.}~\cite{meeran2024sam} utilized the SAM image encoder as a feature extraction backbone and introduced a module to propagate the initial ground truth mask across subsequent frames.
Compared to existing approaches, our method integrates both motion and appearance cues to autonomously generate and refine reliable prompts for SAM2, enabling optimized performance without the need for user-provided prompts.

\subsection{Segment Anything Model}
The Segment Anything Model (SAM)~\cite{kirillov2023segment} has demonstrated remarkable performance in natural image segmentation, particularly due to its robust zero-shot capabilities. However, SAM’s effectiveness can vary significantly across specialized domains~\cite{chen2023sam}. For instance, it faces challenges in segmenting medical images~\cite{huang2024segment} and detecting camouflaged objects~\cite{tang2023can}. To expand SAM’s applicability in medical imaging, approaches like MedSAM~\cite{ma2024segment} and SAM-Adapter~\cite{chen2023sam-adapter} have been developed, integrating domain-specific knowledge to improve performance.
Building on SAM’s success in the image domain, Meta AI Research introduced SAM2~\cite{ravi2024sam2}, a unified architecture designed for both image and video segmentation tasks. 
SAM is limited to image-level segmentation and requires repeated prompts for each frame. SAM2 introduces a memory mechanism that enables temporal propagation and consistency, reducing user interaction while improving efficiency and robustness. This extension makes SAM2 applicable to real-time video segmentation and tracking.
This advancement for SAM2 has prompted further research, with methods such as MedSAM2~\cite{zhu2024medical} and SAM2-Adapter~\cite{chen2024sam2-adapter} integrating specialized knowledge to tailor SAM2 for specific applications. However, to date, no efforts have been made to adapt SAM2 for the VCOD task. 
To bridge this gap, we introduce a novel framework that employs motion- and appearance-guided prompts, alongside an automatic multi-prompts optimization mechanism specifically designed for SAM2. Our proposed framework marks a pioneering application of SAM2 in video camouflaged object detection.

\section{Proposed Method}\label{sec:method}
\textit{Discussion of freezing SAM2:}
The SAM2-Adapter~\cite{chen2024sam2-adapter} is an advanced adaptation strategy for SAM2, yet its performance on VCOD is limited ($S_{\alpha}$ = 0.569, see Table~\ref{tab: Quantitative_Segmentation_Result_VCOD}), far below recent specialized models. This is mainly due to the characteristics of VCOD: (1) the dataset is small and highly redundant, offering insufficient diversity for robust adaptation; and (2) its low-resolution videos blur boundaries and suppress discriminative details, making adaptation ineffective. As a result, SAM2-Adapter fails to generalize well to camouflage scenarios, while directly prompting SAM2 proves more effective~\cite{zhou2025sam2}. Based on the above discussion, our approach chooses to directly freeze SAM2 instead of performing adaptive retraining. Concurrently, rather than simply utilizing SAM2 as an entire entity, we leverage its multi-frame prompting capabilities and the functionalities of each module, designing a novel prompting scheme and skillfully decomposing core modules for utilization to obtain better prediction results.

The overall architecture of our proposed \ourmodel~is illustrated in Fig.~\ref{fig:framework}. The framework is composed of two main components: 
(A) \textbf{Video-based Adaptive Multi-Prompts Refinement} (\ourAlgorithm), which operates through three progressive stages: camouflaged targets determination, pivotal prompt frame selection, and multi-prompts formation. Then these generated prompts are interacted with SAM2 for the final prediction.
(B) \textbf{Motion-Appearance Prompt Inducer} (\ournet), which acts as a camouflaged perception and provides coarse prompts for subsequent \ourAlgorithm~processing.
Each of these components will be detailed in the following sections.

\subsection{Video-based Adaptive Multi-Prompts Refinement}
The quality of prompts directly determines the segmentation performance of SAM2. However, in camouflage scenarios, factors such as cluttered backgrounds, ambiguous boundaries, and scene transitions can significantly undermine the robustness of the masks generated by the prompt inducer. Consequently, effectively filtering out erroneous prompts and optimizing prompt quality emerges as a crucial step toward achieving more robust and accurate segmentation outcomes.

Given that SAM2 supports prompting at arbitrary frames with bi-directional information propagation, it can enhance temporal consistency and robustness in video segmentation. However, in camouflaged scenarios, where objects blend into the background and additional factors such as camera motion, perspective shifts, and occlusion often cause object disappearance in certain frames, frame-wise segmentation becomes highly unreliable. Therefore, exploiting SAM2’s multi-frame prompting to supply richer contextual cues is a promising strategy, while the critical challenge lies in identifying key frames with stronger appearance or motion cues to fully unleash its segmentation potential.
Additionally, refining prompts on these key frames is essential to further enhance their reliability.
To address these challenges, we propose a parameter-free, video-based adaptive multi-prompts refinement method comprising three main steps: 

\begin{algorithm}[t]   
    \caption{The proposed \ourAlgorithm}
    \label{alg:multi_prompt}
    \begin{algorithmic}[1]
        \STATE \textbf{Input:} Video $V = \{I_1, \ldots, I_T\}$; Initial masks $\boldsymbol{y} = \{y_1, \ldots, y_T\}$; Thresholds $\tau, \beta$; Hyperparameters $\alpha, m$
        \STATE \textbf{Output:} Final predictions $\hat{\boldsymbol{y}} = \{\hat{y}_1, \ldots, \hat{y}_T\}$

        \STATE \textbf{Step 1: Camouflaged object determination}
        \STATE Initialize $\mathcal{C} \leftarrow \emptyset$ to store region counts
        \FOR{$t = 1$ to $T$}
            \STATE Binarize mask $y'_t$, apply morphological closing
            \STATE Compute connected regions $R_t$, let $n_t = |R_t|$
            \STATE Update dictionary: $\mathcal{C}(n_t) \leftarrow \mathcal{C}(n_t) + 1$
        \ENDFOR
        \STATE $N_{\max} \leftarrow \arg\max_{n_t} \mathcal{C}(n_t)$
        \STATE Assign IDs to targets in $I_x$ based on $\mathcal{C}(n_t) = N_{\max}$
        \FOR{each subsequent frame $I_t$}
            \FOR{each target ID from $I_{t-1}$}
                \STATE Match targets using IoU ($\text{IoU} > \tau$)
            \ENDFOR
        \ENDFOR
        \STATE Repeat similar steps for previous frames

        \STATE \textbf{Step 2: Pivotal prompt frame selection}
        \FOR{each target ID}
            \FOR{each frame $I_t$}
                \STATE Input $m$ prompt points $P_t$ into SAM2 to obtain $y_t''$
            \ENDFOR
        \ENDFOR
        \STATE Select top $k$ frames: $I_{\text{selected}} = \text{top-}k\left(\text{IoU}(y^{\prime}_t, y^{\prime\prime}_t)\right)$

        \STATE \textbf{Step 3: Multi-prompts formation}
        \FOR{each selected frame $I_t$}
            \STATE Generate initial box $B_t$ around $y_t''$
            \FOR{each direction (up, down, left, right)}
                \REPEAT
                    \STATE Expand box $B_t$ with step size $\alpha$
                    \STATE Compute mask change $\Delta M$
                \UNTIL{$\Delta M \geq \beta$}
            \ENDFOR
            \STATE Save final box $B_t$
        \ENDFOR

        \STATE Input $\mathbb{B} = \{B_a, \dots, B_n \}$, $\mathbb{P} = \{P_a, \dots, P_n \}$, and $I_{\text{selected}} = \{I_a, \dots, I_n \}$ into SAM2 to get final predictions: 
        $\hat{\boldsymbol{y}}$ = \text{SAM2}($\mathbb{B}, \mathbb{P}, I_{\text{selected}}$)
        
    \end{algorithmic}
\end{algorithm}

\textbf{Step 1: Camouflaged object determination.}  
To address boundary ambiguity caused by camouflaged objects blending with backgrounds, we refine initial masks $\boldsymbol{y}$ through a two-step process: (1) Binarization removes low-confidence edge pixels (pixel value $\leq \tau$) to suppress noise and emphasize high-confidence regions near the object center ($\tau$ is set to 127). (2) Morphological closing eliminates residual background artifacts and pseudo-targets, yielding refined masks \( \boldsymbol{y}^\prime \).
We then estimate target count by analyzing the connected-region frequency across frames. Notably, for scenarios that are estimated to be single target, a top-$k$ frame selection filters out multi-region noisy frames (which inherently exhibit low similarity in Step 2), preventing their selection as prompt frames. Multi-target cases trigger our ID assignment protocol (Algorithm~\ref{alg:multi_prompt}).

\textbf{Step 2: Pivotal prompt frame selection.}
For SAM2, point prompts offer a practical and user-friendly approach, eliminating the need for specialized knowledge and facilitating real-world applicability.
We treat the prompt inducer and SAM2 as two complementary experts and adopt a bi-directional decision scheme in which both experts jointly select high-quality frames—\ie, frames where the camouflaged object is easier to segment and the predicted masks are stronger. Such frames carry richer object evidence; using them as prompts seeds SAM2 with more informative representations, promotes more effective temporal propagation, and ultimately improves camouflaged object segmentation.

As illustrated in Fig.~\ref{fig:framework}, we first generate random prompt points from $\boldsymbol{y}^\prime$ and input these points, together with the corresponding RGB frames, into SAM2 to obtain single-frame predictions $\boldsymbol{y}^{\prime\prime}$. 
Subsequently, each prediction in $\boldsymbol{y}^{\prime\prime}$ is compared with its counterpart in $\boldsymbol{y}^\prime$ to measure similarity.
For computational efficiency, we adopt Intersection over Union (IoU) as the similarity metric. This comparison is conducted across all frames within the video clip, enabling us to rank them by similarity and select the top-$k$ frames with the highest scores.

\textit{Implementation note:} features extracted by SAM2 are buffered in a newly designed feature memory, which is distinct from SAM2’s internal memory bank used for temporal propagation. Subsequent operations read directly from this feature memory rather than re-extracting features, thereby reducing overall computation.
 
\textbf{Step 3: Multi-prompts formation.} 
As demonstrated in work~\cite{zhou2025sam2}, mask prompts can achieve highly precise segmentation in VCOD by directly specifying the target region’s shape, thereby minimizing errors. They are particularly effective for complex or irregular objects, as they capture fine details such as holes, textures, or subtle boundaries. However, mask prompts are inherently inflexible; even small inaccuracies (e.g., missing a minor part) can propagate to the final segmentation. Without manual intervention, generating precise camouflage masks is nearly impossible under current conditions.
In contrast, point prompts are simple and intuitive, requiring minimal input while effectively highlighting small or distinct objects. Box prompts provide stronger spatial constraints, thereby reducing the ambiguity of point-only prompts; yet overly broad boxes may include irrelevant background pixels, leading to over-segmentation. Therefore, we propose combining point and box prompts: the box defines the overall object scope, while points refine local details, achieving both coverage and precision. 

Specifically, for the selected top-$k$ frames, we refine prompt boxes to address under-segmentation caused by point prompts’ tendency to focus on local regions.
Initially, prompt boxes are generated from SAM2’s segmentation masks $\boldsymbol{y}^{\prime\prime}$ using randomly chosen points, with the boxes initialized as the minimum enclosing rectangles around these masks. 
Given that the initial prediction results $\boldsymbol{y}^{\prime\prime}$ are typically smaller than the ground truth (GT) due to the point prompt potentially guiding the model to segment only local regions rather than the entire target object, each box is expanded outward in four directions—up, down, left, and right. The expansion process proceeds in each direction until a significant change is observed, which then serves as the stopping criterion for that direction.
This process is repeated for all four directions until the final bounding box is established. 
Finally, the optimized boxes, along with their respective prompt points and frames, are then inputted back into SAM2 to produce final predictions: $\hat{\boldsymbol{y}}=\{\hat{y}_{1},\hat{y}_{2},...,\hat{y}_{T}$\}.
The visual optimization process for the pivotal frame $i$, transitioning from $y_i^{\prime\prime}$ to $\hat{y_i}$ is illustrated in Fig.~\ref{fig:enlarge box}. This progression involves transforming ``Point" into ``Point+box", and subsequently into ``Point+rbox".

\begin{figure*}[t!]
\begin{center}
\includegraphics[width=1.0\textwidth]{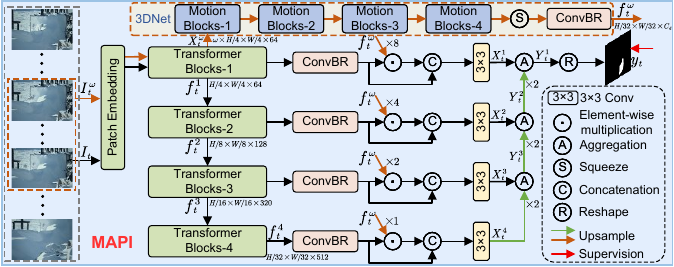}
\end{center}
\vspace{-8pt}
\caption{
Overall architecture of the proposed MAPI. The model begins with a pretrained transformer backbone to extract multi-scale appearance features from the input image. Subsequently, a $\mathtt{3DNet}$ module is employed to capture temporal motion relationships across preceding frames. Finally, a progressive coarse-to-fine decoder, guided by motion-informed appearance features, is applied to progressively refine the segmentation predictions.}
\label{figure_MAPI}
\end{figure*}

\subsection{Motion-Appearance Prompt Inducer}
\textit{Discussion :} 
The prompt inducer provides camouflage perception and initial localization, while SAM2 leverages its strong segmentation capability to refine these coarse masks. Accurate localization is more critical than precise boundaries: when localization is correct, AMPR can adjust confidence regions and optimize prompts to yield significant improvements; when it fails, refinement cannot recover performance, leading to errors on non-target regions. Thus, the performance gain from coarse to refined masks reflects the inducer’s localization ability, which jointly determines camouflage perception and segmentation quality. 
Localization ability can be gauged by the improvement from coarse to AMPR+SAM2-refined masks. As evidenced in Table~\ref{tab: Model+AMPR} and Fig.~\ref{fig:Sa-change}, many existing models achieve good segmentation quality yet remain limited by poor localization. This motivates our motion-guided prompt inducer, which enhances localization while maintaining strong camouflage perception.
Notably, the inducer is a modular component that can be instantiated by existing VCOD networks, as further analyzed in Sec.~\ref{sec:ablation}.

In video-based tasks, both temporal-spatial relationships and appearance cues play equally critical roles in effective target detection.
This underscores the importance of motion-appearance induced prompts specifically designed for SAM2 to enhance its effectiveness in video camouflaged object detection. Therefore, we design a new model MAPI, which is composed of three key parts: appearance feature extraction, motion perception, and motion-guided appearance decoder.

\subsubsection{Appearance Feature Extraction}
Appearance information encompasses the visual characteristics of an object, including color, texture, and shape. This information is crucial for discerning the subtle differences between camouflaged objects and their backgrounds. 
Vision transformer-based models \cite{cheng2022implicit,huang2023feature,hui2024implict-explicit,hui2024endow,zhang2025emip,pang2022zoom,fan2021concealed} have demonstrated impressive capabilities in modeling both global and local contexts for the task of detecting camouflaged objects in images.
Notably, the design of the appearance feature extraction network is not the primary focus of this paper; therefore, for fair comparisons, we employ the PVT~\cite{wang2022pvt} as our feature extraction backbone, adhering to the same configurations as those in \cite{cheng2022implicit,hui2024endow} without additional modifications. Specifically, for a given reference frame \( I_t \), we extract a set of features $\{f_{t}^{i} \in \mathbb{R}^{H/{2^{i+1}} \times W/{2^{i+1}}\times C_i}, i=1, \ldots, 4\}$ at varying scales from PVT. Here, $W$, $H$, and $C$ denote the width, height, and channel number, respectively.

\subsubsection{Motion Perception}
Camouflaged objects closely resemble their surroundings, making them difficult to detect, even for the human eye. However, any movement from the target can disrupt this concealment, revealing its presence. Leveraging this, we incorporate motion information to enhance localization.

As shown in Fig.~\ref{figure_MAPI}, for a video sequence $I^{\omega}_{t} \in \mathbb{R}^{\omega \times H \times W \times 3}$ of window length $\omega$ corresponding to the current frame $I_t$, we first feed it into the initial block of PVT backbone to obtain the low-level feature map $X^{\omega}_{t} \in \mathbb{R}^{\omega \times H/4 \times W/4 \times 64}$. Then we apply a 3D convolutional neural network ($\mathtt{3DNet}$) to capture the temporal-spatial relationships across frames:
\begin{equation}
    {f}^{\omega}_t=\mathtt{3DNet}(X^{\omega}_{t}),
\end{equation}
where ${f}^{\omega}_t \in \mathbb{R}^{1 \times H/32 \times W/32 \times 512}$ represents the temporal-spatial relationships within $\omega$ consecutive frames. 
Referring to the long-term setting of SLT-Net~\cite{cheng2022implicit}, we set $\omega$ to 5 in this paper.
The $\mathtt{3DNet}$ consists of four sequential motion blocks, a squeeze operation, and a $\mathtt{ConvBR}$ module. 
Specifically, each motion block operates as follows:
\begin{equation}
    y=\phi(x+\mathtt{Down}(\mathtt{BN}(\mathtt{Conv3D}(x)))),
\end{equation}
where $x$ and $y$ denote the input and output, respectively.
$\phi$ denotes the ReLU activation function~\cite{hendrycks2016gaussian}, and $\mathtt{BN}$ refers to batch normalization.
The operator $\mathtt{Conv3D}$ represents a 3$\times$3$\times$3 convolution with a stride of 2$\times$2$\times$2, while $\mathtt{Down}$ denotes a 1$\times$1$\times$1 convolution with a stride of 2$\times$2$\times$2, followed by batch normalization.
And the $\mathtt{ConvBR}$ module is composed of a 3$\times$3 convolutional layer, followed by batch normalization and a ReLU activation. It is specifically employed to reduce the channel dimensions of $f_t^{\omega}$ to 64 channels.
In particular, to preserve the spatial resolution of the final output and ensure alignment with the feature tensor $f_{t}^{4}$, the downsampling operation ($\mathtt{Down}$) is omitted in the first motion block.

\subsubsection{Motion-guided Appearance Decoder}
To effectively integrate motion and appearance information for obtaining robust masks, we employ a multi-stage fusion and progressive decoding strategy, as illustrated in Fig.~\ref{figure_MAPI}. 
Initially, we apply four independent $\mathtt{ConvBR}$ modules to reduce the channel dimensions of features \(f_t^{\omega}, f_t^{2}, f_t^{3}, f_t^{4}\) to 64 channels.
Next, the motion feature \(f_t^{\omega}\) is resized to match the spatial resolution of the corresponding appearance feature map \(f_t^{2}, f_t^{3}, f_t^{4}\). These features are then fused through element-wise multiplication and concatenation, followed by a \(3 \times 3\) convolutional layer.
To address the challenges of detecting small objects, which often lose clarity in down-sampled feature maps, and recognizing that motion features are primarily beneficial for localization, we establish feature interactions that progress from coarse to fine resolutions. 
The aggregated features $\{Y_i\}^3_{i=1}$ can be wirtten as:
\begin{equation}
    Y_i = \mathtt{ConvBR}([X^{i}_t,\mathcal{F}_{\mathtt{up}}(Y^{i+1}_t)]),
\end{equation}
where $[\cdots]$ denotes the concatenation operation, and $\mathcal{F}_{\mathtt{up}}(\cdot)$ represents a bilinear upsampling operation used for spatial resolution alignment. Specifically, at the initial stage of aggregation, the input feature $Y^{4}_t$ is directly initialized as $X^{4}_t$. Finally, the final feature map is reshaped to the same size as the input image.

\subsection{Supervision and Loss Function}
Following previous methods~\cite{cheng2022implicit,zhang2025emip}, we perform joint optimization for the prompt inducer with both motion and appearance cues by minimizing a hybrid loss function~\cite{21Fan_HybridLoss}, defined as follows:
\begin{equation}
    \mathcal{L}_{\text{pred}} = \omega_1\mathcal{L}_{\text{IoU}} + \omega_2\mathcal{L}_{\text{bce}} + \omega_3\mathcal{L}_{\text{e-loss}}, 
\end{equation}
where $\mathcal{L}_{\text{IoU}},\mathcal{L}_{\text{bce}}$, and $\mathcal{L}_{\text{e-loss}}$ denote IoU loss, binary cross-entropy loss, and enhanced-alignment loss, respectively.
Note that all weight coefficients of the loss terms are set to 1 (\ie, $\omega_1=\omega_2=\omega_3$), which is equivalent to equally weighting each loss term during model training.

\section{Experiment Results and Analyses}\label{sec:exp}
\subsection{Datasets and Evaluation Metrics}
\noindent\textbf{Datasets.} 
Following previous VCOD methods~\cite{cheng2022implicit,hui2024implict-explicit,hui2024endow,meeran2024sam}, we conduct experiments on two widely recognized VCOD benchmarks: MoCA-Mask~\cite{cheng2022implicit} and CAD~\cite{bideau2016s}. MoCA-Mask is recognized as the more challenging dataset, featuring camouflaged animals in natural environments. It consists of 19,313 frames derived from 71 video clips for training and 3,626 frames from 16 clips for testing. Conversely, the CAD dataset is a smaller collection specifically for testing, comprising 836 frames from 9 clips sourced from YouTube videos.

\noindent\textbf{Evaluation metrics.}
We adopt widely recognized evaluation metrics to assess our model performance, namely: structure measure ($\mathcal{S}_\alpha$)~\cite{fan2017structure}, weighted F-measure ($F_{\beta}^{w}$)~\cite{margolin2014evaluate}, 
enhanced-alignment measure ($E_{\phi}$~\cite{fan2018enhanced}),
mean absolute error ($\mathcal{M}$)~\cite{perazzi2012saliency}, and mean value of Dice (mDice) and IoU (mIoU). These metrics provide a comprehensive and reliable assessment of model performance.

\subsection{Implementation Details}
Our \ourmodel~is implemented by PyTorch~\cite{paszke2019pytorch} on a single NVIDIA 4090 GPU and optimized with Adam optimizer by cosine annealing strategy, whose maximum, minimum learning rates, and the maximum adjusted iteration are set to 1e-5, 1e-6, and 20, respectively. 
The parameters $\tau,\alpha,\beta, m$ in Algorithm~\ref{alg:multi_prompt} are set to 0.5, 5, 5e-4, and 5, respectively.  
For fair comparisons, we strictly follow the training configurations detailed in \cite{cheng2022implicit,hui2024endow}, employing PVTv2~\cite{wang2022pvt} as the feature extraction backbone. And the PVTv2 backbone is also pre-trained on the static training set of COD10K (3,040 images)~\cite{fan2021concealed}. All input images are resized to 352$\times$352. The model is trained on the training set of MoCA-Mask (19,313 frames)~\cite{cheng2022implicit} and evaluated on the MoCA-Mask test set, as well as on the entire CAD dataset. 
The mini-batch is set to 6, and MAPI trains for 4 hours over 60 epochs. 
We choose the hiera-small version of SAM2 in all our experiments.
To ensure robust results, all experiments were conducted five times, with the median result reported in the following tables. 

\subsection{Quantitative and Qualitative Comparison}\label{sec:Comparison with State-of-the-Arts}
\begin{table*}[t!]
\centering
\caption{Quantitative comparisons on MoCA-Mask and CAD datasets. The top three results are highlighted in \textcolor{red}{red}, \textcolor{indiagreen}{green}, and \textcolor{blue}{blue}.}
\tabcolsep=0.1cm
\resizebox{\textwidth}{!}{
\begin{tabular}{l|c|c|cccccc|cccccc}
\toprule
\multirow{2}{*}{Model} & \multirow{2}{*}{Publication} & \multirow{2}{*}{Input} & \multicolumn{6}{c|}{MoCA-Mask} & \multicolumn{6}{c}{CAD} \\ \cline{4-15} 
 &  &  & $S_\alpha \uparrow$ & $F_{\beta}^{w} \uparrow$ & $E_{\phi} \uparrow$ & $\mathcal{M} \downarrow$ & mDice$\uparrow$ & mIoU$\uparrow$ & $S_\alpha \uparrow$ & $F_{\beta}^{w} \uparrow$ & $E_{\phi} \uparrow$ & $\mathcal{M} \downarrow$ & mDice$\uparrow$ & mIoU$\uparrow$ \\ 
\midrule
EGNet~\cite{zhao2019EGNet} & ICCV-2019 & Image & 0.574 & 0.110 & 0.574 & 0.035 & 0.143 & 0.096 & 0.619 & 0.298 & 0.666 & 0.044 & 0.324 & 0.243 \\
BASNet~\cite{Qin_2019_CVPR} & CVPR-2019 & Image & 0.561 & 0.154 & 0.598 & 0.042 & 0.190 & 0.137 & 0.639 & 0.349 & 0.773 & 0.054 & 0.393 & 0.293 \\
PraNet~\cite{fan2020pra} & MICCAI-2020 & Image & 0.614 & 0.266 & 0.674 & 0.030 & 0.311 & 0.234 & 0.629 & 0.352 & 0.763 & 0.042 & 0.378 & 0.290 \\
SINet~\cite{fan2020Camouflage} & CVPR-2020 & Image & 0.574 & 0.185 & 0.655 & 0.030 & 0.221 & 0.156 & 0.601 & 0.204 & 0.589 & 0.089 & 0.289 & 0.209 \\
SINet-v2~\cite{fan2021concealed} & TPAMI-2021 & Image & 0.571 & 0.175 & 0.608 & 0.035 & 0.211 & 0.153 & 0.544 & 0.181 & 0.546 & 0.049 & 0.170 & 0.110 \\
ZoomNet~\cite{pang2022zoom} & CVPR-2022 & Image & 0.582 & 0.201 & 0.682 & 0.026 & 0.236 & 0.197 & 0.661 & 0.235 & 0.666 & 0.089 & 0.345 & 0.265 \\
BGNet~\cite{sun2022boundary} & IJCAI-2022 & Image & 0.590 & 0.203 & 0.647 & 0.023 & 0.225 & 0.168 & 0.651 & 0.240 & 0.625 & 0.077 & 0.320 & 0.238 \\
FEDERNet~\cite{he2023FEDER} & CVPR-2023 & Image & 0.555 & 0.198 & 0.542 & 0.049 & 0.192 & 0.152 & 0.604 & 0.233 & 0.725 & 0.061 & 0.361 & 0.301 \\
FSPNet~\cite{huang2023feature} & CVPR-2023 & Image & 0.565 & 0.186 & 0.610 & 0.044 & 0.238 & 0.167 & 0.609 & 0.224 & 0.664 & 0.056 & 0.315 & 0.235 \\
SAM2-adapter~\cite{chen2024sam2-adapter} & arXiv-2024 & Image & 0.569 & 0.162 & 0.586 & 0.041 & 0.213 & 0.144 & 0.650 & 0.387 & 0.746 & 0.043 & 0.442 & 0.329 \\
FSEL~\cite{sun2024frequency} & ECCV-2024 & Image & 0.596 & 0.260 & 0.677 & 0.053 & 0.219 & 0.151 & 0.649 & 0.368 & 0.732 & 0.053 & 0.434 & 0.325 \\ 

\midrule
RCRNet~\cite{yan2019semi} & ICCV-2019 & Video & 0.555 & 0.138 & 0.527 & 0.033 & 0.171 & 0.116 & 0.627 & 0.287 & 0.666 & 0.048 & 0.309 & 0.229 \\
PNS-Net~\cite{ji2021progressively} & MICCAI-2021 & Video & 0.576 & 0.134 & 0.562 & 0.038 & 0.189 & 0.133 & 0.678 & 0.369 & 0.720 & 0.043 & 0.409 & 0.308 \\
MG~\cite{yang2021selfsupervised} & ICCV-2021 & Video & 0.530 & 0.168 & 0.561 & 0.067 & 0.181 & 0.127 & 0.594 & 0.336 & 0.691 & 0.059 & 0.368 & 0.268 \\
SLT-Net~\cite{cheng2022implicit} & CVPR-2022 & Video & 0.631 & 0.311 & 0.759 & 0.027 & 0.360 & 0.272 & 0.696 & 0.481 & \textcolor{blue}{0.845} & 0.030 & 0.493 & 0.401 \\ 
IMEX~\cite{hui2024implict-explicit} & TMM-2024 & Video & 0.661 & 0.371 & 0.778 & 0.020 & 0.409 & 0.319 & 0.684 & 0.452 & 0.813 & 0.033 & 0.469 & 0.370  \\
TSP-SAM~\cite{hui2024endow} & CVPR-2024 & Video & \textcolor{blue}{0.689} & \textcolor{blue}{0.444} & \textcolor{blue}{0.808} & \textcolor{blue}{0.008} & \textcolor{blue}{0.458} & \textcolor{blue}{0.388} & 0.704 & \textcolor{blue}{0.524} & \textcolor{red}{0.912} & \textcolor{blue}{0.028} & \textcolor{blue}{0.543} & \textcolor{blue}{0.438} \\ 
SAM-PM~\cite{meeran2024sam} & CVPR-2024 & Video & \textcolor{indiagreen}{0.728} & \textcolor{indiagreen}{0.567} & \textcolor{indiagreen}{0.813} & \textcolor{indiagreen}{0.009} & \textcolor{indiagreen}{0.594} & \textcolor{indiagreen}{0.502} & \textcolor{indiagreen}{0.729} & \textcolor{indiagreen}{0.602} & 0.746 & \textcolor{indiagreen}{0.018} & \textcolor{indiagreen}{0.594} & \textcolor{indiagreen}{0.493} \\ 
EMIP~\cite{zhang2025emip} & TIP-2025 & Video & 0.669 & 0.374 & 0.785 & 0.017 & 0.424 & 0.326 & \textcolor{blue}{0.710} & 0.504 & 0.832 & 0.029 & 0.528 & 0.415  \\
\midrule
\textbf{\ournet} & Ours & Video & 0.670 & 0.376 & 0.782 & 0.014 & 0.413 & 0.330 & 0.709 & 0.513 & 0.824 & 0.029 & 0.530 & 0.413 \\
\textbf{\ourmodel} & Ours & Video & \textcolor{red}{0.765} & \textcolor{red}{0.607} & \textcolor{red}{0.848} & \textcolor{red}{0.007} & \textcolor{red}{0.620} & \textcolor{red}{0.542} & \textcolor{red}{0.774} & \textcolor{red}{0.652} & \textcolor{indiagreen}{0.852} & \textcolor{red}{0.018} & \textcolor{red}{0.647} & \textcolor{red}{0.543} \\ 
\bottomrule
\end{tabular}
}
\label{tab: comparison with sota}
\end{table*}

\begin{figure*}[t!]
\begin{center}
\includegraphics[width=1.0\textwidth]{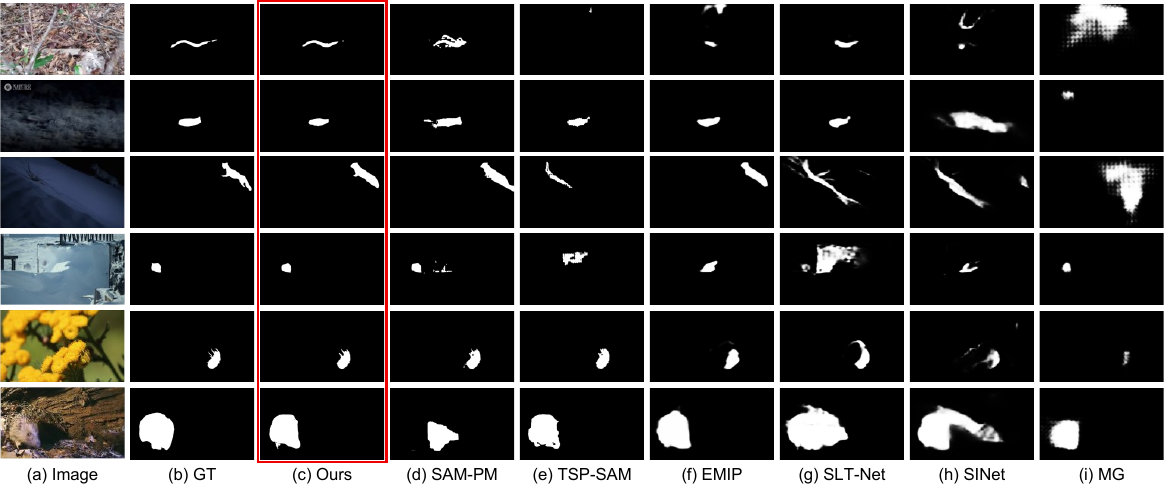}
\end{center}
\vspace{-8pt}
\caption{Visualization of our proposed \ourmodel~and previous state-of-the-art methods on MoCA-Mask and CAD datasets.}
\label{fig:visualization with sota}
\end{figure*}

To evaluate the effectiveness of the proposed \ourmodel, we compare it against a range of state-of-the-art methods, including both image-based and video-based approaches. As shown in Table~\ref{tab: comparison with sota}, the results reveal several key insights: (i) The substantial performance gap between video-based and single-image camouflaged object detection methods highlights the critical role of temporal-spatial relationships in resolving video camouflaged challenges. 
(ii) SAM-based methods achieve superior performance relative to other approaches, underscoring the powerful feature extraction and generalization capabilities of the foundation model. 
(iii) Our motion-appearance prompt inducer outperforms all existing non-SAM-based models in predicting camouflaged objects, showcasing the effectiveness of our simple yet powerful design in extracting and integrating both motion and appearance features.
(\textrm{iv}) The proposed \ourmodel~outperforms all video-based camouflaged object detection methods. 
Notably, it achieves a 6.2\% improvement in $S_\alpha$ on the CAD dataset over the previous state-of-the-art SAM-PM~\cite{meeran2024sam}, suggesting that our \ourmodel~exhibits enhanced robustness and generalization on unseen dataset.
Furthermore, visual comparisons in Fig.~\ref{fig:visualization with sota} show that our \ourmodel~more accurately localizes and segments camouflaged targets compared to other leading methods.

Additionally, we present the model parameter count and frames-per-second (FPS). As detailed in Table \ref{tab: param_comparison}, our model achieves a 0.26 improvement in mDice with only a 59.18M increase in parameters over SLT-Net. 
Notably, our model also achieves the highest FPS among existing VCOD models, underscoring its efficiency and effectiveness.
To further validate the generalizability of our \ourAlgorithm, we apply it to existing VCOD methods. The variants of these models show improved performance compared to their original versions (See details in Sec.~\ref{sec:ablation}). This demonstrates that the incorporation of \ourAlgorithm~can substantially enhance the performance of a VCOD model, even when the baseline model is not particularly strong. Furthermore, when applied to a more robust model, such as our \ournet, the \ourAlgorithm~generates even better results, underscoring the close interdependence between these two designed components in our method.

\begin{table}[h!]
    \footnotesize
    \centering
    \caption{Comparison of model parameters and FPS with state-of-the-art methods. The best scores are highlighted in \textbf{bold}.}
    \tabcolsep=0.12cm
    \begin{tabular}{c|ccccccc}
    \toprule
      Model & Params & FPS & $\mathcal{S}_\alpha\uparrow$ &$F_\beta^w\uparrow$ & $E_{\phi} \uparrow$ & mDice$\uparrow$\\
     \midrule
     FSPNet~\cite{huang2023feature} & 274.24M & 2.41 & 0.565 & 0.186 & 0.610 & 0.238 \\
     SLT-Net~\cite{cheng2022implicit} & \textbf{82.38M} & 5.52 & 0.631 & 0.311 & 0.759 & 0.360 \\
     TSP-SAM~\cite{hui2024endow} & 727.12M & 2.69 & 0.689 & 0.044 & 0.808 & 0.458 \\
     SAM-PM~\cite{meeran2024sam} & 313.33M & 5.08 & 0.728 & 0.567 & 0.813 & 0.594 \\
     \ourmodel~(ours) & 141.56M & \textbf{6.78} & \textbf{0.765} & \textbf{0.607} & \textbf{0.848} & \textbf{0.620} \\
    \bottomrule
    \end{tabular}
    \label{tab: param_comparison}
\end{table}

\subsection{Ablation Studies}\label{sec:ablation}
To comprehensively assess the effectiveness of the key components and the selection of hyperparameters, we perform an in-depth analysis by decoupling the core design and varying hyperparameter values.

\begin{table}[h!]
\centering
\caption{Ablation studies of the core components of our proposed \ournet~on MoCA-Mask dataset.}
\footnotesize
\renewcommand{\arraystretch}{1.1}
\renewcommand{\tabcolsep}{0.10cm}
\begin{tabular}{c|cc|cccccc}
    \toprule
    \# & Appearance & Motion &
    $\mathcal{S}_\alpha\uparrow$ &$F_\beta^w\uparrow$ &
    $E_{\phi} \uparrow$ &
    $\mathcal{M}\downarrow$ & mDice$\uparrow$ & mIoU$\uparrow$ \\
    \midrule
    1 & - & - & 0.624 & 0.289 & 0.781 &0.024 & 0.337 & 0.251 \\
    2 & \Checkmark & - & 0.631 & 0.296 & 0.764 & 0.024 & 0.341 & 0.258  \\
     3 & \Checkmark & \Checkmark
     & \textbf{0.670} & \textbf{0.376} & \textbf{0.782} & \textbf{0.014} & \textbf{0.413} & \textbf{0.330}\\
    \bottomrule
\end{tabular}
\label{tab: motion_appearance_ablation}
\end{table}

\noindent\textbf{Ablation analysis of \ournet.}
Table~\ref{tab: motion_appearance_ablation} evaluates the segmentation results of \ournet~by progressively incorporating each module. The ``Baseline" (\#1) refers to using only the final layer of the PVT feature extraction backbone for predictions. The ``Appearance" (\#2) configuration decodes multi-layer feature maps from different stages of the backbone to generate predictions. As observed, decoding multi-stage appearance features significantly enhances performance. However, when motion information is introduced (\#3, our \ournet) with $\mathtt{3DNet}$, the mDice metric notably improves from 0.341 to 0.413, demonstrating the critical role of inter-frame motion addressed by our $\mathtt{3DNet}$ in breaking camouflage. Additionally, in comparison to the results in Table~\ref{tab: comparison with sota}, \ournet~outperforms all previous non-SAM-based methods, further underscoring the effectiveness of our design in video camouflaged object perception.

\noindent\textbf{Prompt frame selection.}
Table~\ref{tab: top_strategy_validation} evaluates the impact of different frame selection schemes on MoCA-Mask dataset. Three prompt selection strategies are compared: using the first frame of a video sequence (``first"), a randomly chosen frame (``random"), and the frame selected by our proposed \ourAlgorithm~(``top-1"), with each prompted frame accompanied by a single random prompt point.
As shown, the ``top-1" selection strategy outperforms both the ``first" and ``random" selection strategies across all evaluation metrics, achieving substantial improvements of 0.233 and 0.201 in mDice, respectively.

To further validate the effectiveness of our prompt frame selection strategy and control for other confounding factors, we replace the coarse predictions generated by \ournet~with user-provided masks, thereby eliminating prompt-induced errors introduced by the initial network. Then, we evaluate the impact of the prompt frame location on segmentation performance using a 1-click prompt. For simplicity, we replace the user-provided mask with the ground truth mask and generate prompt points accordingly. 
As shown in Table~\ref{tab: gt_frame_selection}, the ``top-1" frame consistently achieves the highest segmentation performance across all evaluation metrics, while using the middle frame yields the second-best results. 
This suggests that, although the middle frame often provides more informative cues compared to randomly selected or boundary-adjacent frames—likely due to its temporal centrality and relatively stable appearance—the ``top-1'' frame identified by our method offers more effective guidance. 
These findings confirm that the choice of prompt frame plays a crucial role in improving SAM2’s segmentation quality.
Our targeted selection strategy is better suited for adaptation to downstream tasks.

\begin{table}[t!]
    \footnotesize
    \centering
    \caption{Comparing different selection methods within our model.}
    \tabcolsep=0.21cm
    \begin{tabular}{c|cccccc}
    \toprule
     Frame& $\mathcal{S}_\alpha\uparrow$ &$F_\beta^w\uparrow$ & $E_{\phi} \uparrow$ & $\mathcal{M}\downarrow$ & mDice$\uparrow$ & mIoU$\uparrow$ \\
     \midrule
     first & 0.627 & 0.330 & 0.703 & 0.046 & 0.337 & 0.294\\
     random & 0.629 & 0.363 & 0.675 & 0.065 & 0.369 & 0.324\\
     top-1 & \textbf{0.745} & \textbf{0.561} & \textbf{0.805} & \textbf{0.008} & \textbf{0.570} & \textbf{0.501}\\
    \bottomrule
    \end{tabular}
    \label{tab: top_strategy_validation}
\end{table}

\begin{table}[htbp]
\centering
\caption{Performance of SAM2 with different prompt frame.}
\tabcolsep=0.070cm
\begin{tabular}{c|c|cccccc}
\toprule
~~~\multirow{2}{*}{\textbf{Prompt}}~~~ & ~~~\multirow{2}{*}{\textbf{Frame}}~~~ & \multicolumn{6}{c}{\textbf{MoCA-Mask} \cite{cheng2022implicit}} \\ \cline{3-8} 
 &  & $S_m \uparrow$ & $F_\beta^\omega \uparrow$ & $E_{\phi} \uparrow$ & $\mathcal{M}\downarrow$  & mDice $\uparrow$ & mIoU $\uparrow$ \\ 
\midrule
\multirow{9}{*}{1-click} 
 & 0 & 0.710 & 0.536 & 0.737 & 0.009 & 0.537 & 0.436 \\
 & 5 & 0.705 & 0.510 & 0.718 & 0.054 & 0.513 & 0.444  \\
 & 10 & 0.680 & 0.520 & 0.713 & 0.070 & 0.520 & 0.443 \\
 & -11 & 0.720 & 0.541 & 0.771 & 0.056 & 0.554 & 0.486 \\
 & -6 & 0.710 & 0.593 & 0.744 & 0.115 & 0.599 & 0.519 \\
 & -1 & 0.689 & 0.510 & 0.723 & 0.053 & 0.511 & 0.443  \\
 & middle & 0.733 & 0.551 & 0.749 & 0.007 & 0.555 & 0.475 \\ 
 & random & 0.709 & 0.528 & 0.730 & 0.052 & 0.532 & 0.456  \\
 & top-1 & \textbf{0.804} & \textbf{0.690} & \textbf{0.855} & \textbf{0.005} & \textbf{0.693} & \textbf{0.605}  \\
 \bottomrule
\end{tabular}%
\label{tab: gt_frame_selection} 
\end{table}

\begin{figure}[h]
\begin{center}
\includegraphics[width=.478\textwidth]{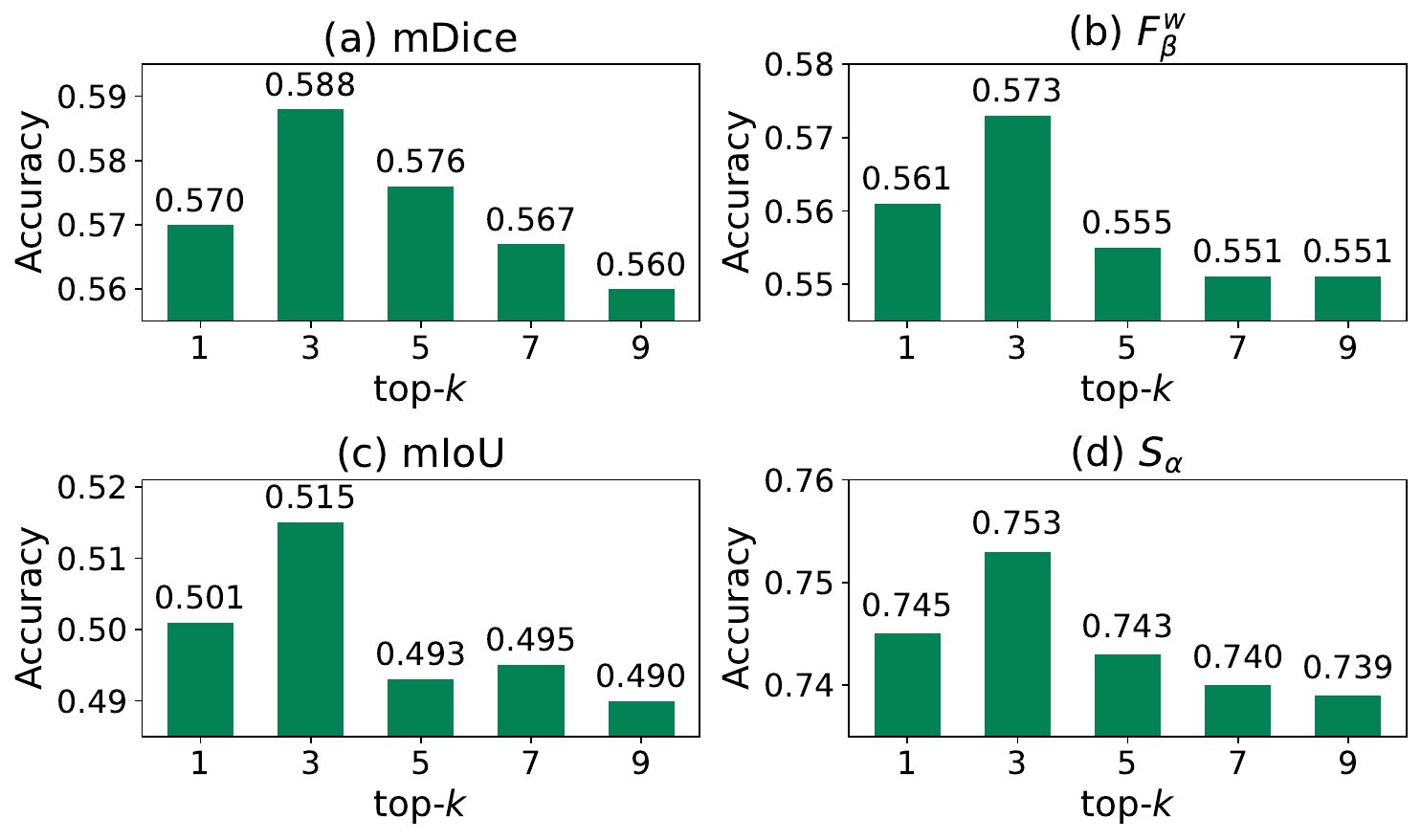}
\end{center}
\vspace{-8pt}
\caption{Sensitivity analysis of the number of top-$k$ prompt frames on MoCA-Mask dataset.}
\label{fig:metrich_vs_top-k}
\end{figure}

\noindent\textbf{Hyper parameter analysis.}
To evaluate the impact of the number of prompt frames on segmentation performance, we conduct ablation studies by varying $k$ within the set $\{1, 3, 5, 7, 9\}$. As shown in Fig.~\ref{fig:metrich_vs_top-k}, the performance improves significantly from top-1 to top-3, indicating that incorporating a small number of additional frames enhances SAM2's ability to generalize the appearance and spatial context of camouflaged objects, thereby leading to more accurate segmentation. However, beyond top-3, performance begins to decline. This suggests that an excessive number of prompt frames may introduce noise or irrelevant information—particularly problematic when the initial masks produced by the prompt inducer are coarse or inaccurate.

In addition, we investigate the sensitivity to the number of prompt points (denoted as $n$p), with value $n$ ranging from 1 to 9, as reported in Table~\ref{tab: rbbox}. The results show that segmentation accuracy increases with more prompt points and peaks at five points, after which performance begins to drop. This trend can be attributed to the ambiguous boundaries and imprecise localization of initial prompts. Using too many points (\eg, 7p and 9p) increases the risk of including noisy or irrelevant regions, such as background clutter or other nearby objects. Conversely, using too few points (\eg, 1p and 3p) provides insufficient spatial and contextual cues, thereby limiting the model's segmentation capability.

\noindent\textbf{Effectiveness of refined box prompt.}
We investigate the impact of mixed prompts on segmentation performance. As shown in the last two rows of Table~\ref{tab: rbbox}, simply introducing diverse prompt types does not necessarily lead to improved performance. Specifically, incorporating minimum bounding box prompts directly derived from the initial masks, alongside point prompts, results in a performance decline. This degradation is primarily attributed to the coarse and often inaccurate nature of the initial bounding boxes, which tend to cause under-segmentation and misguide the model.
In contrast, when refined bounding boxes generated by \ourAlgorithm~are employed, the model achieves an approximate 0.122 improvement in mDice. This substantial gain underscores the importance of box prompt quality: high-precision boxes, when used in conjunction with point prompts, significantly enhance segmentation outcomes. These findings highlight the model’s sensitivity to prompt accuracy and the critical role of precise spatial guidance in optimizing performance.

\begin{figure*}[t!]
\begin{center}
\includegraphics[width=1.0\textwidth]{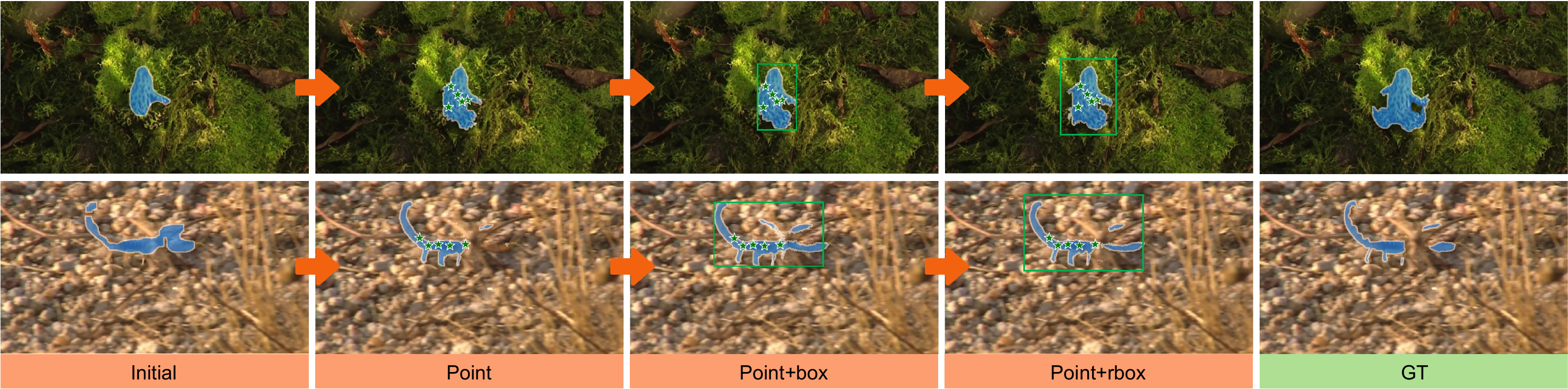}
\end{center}
\vspace{-8pt}
\caption{Visualization of the proposed adaptive multi-prompt refinement process. ``Initial'' denotes the coarse masks obtained after preprocessing in Step 1 of Algorithm~\ref{alg:multi_prompt}. It can be observed that the refinement progressively improves the segmentation quality: from ``Point'' to ``Point+rbox'', the camouflaged objects become more complete with richer details, while noise and irrelevant regions are further suppressed.}
\label{fig:enlarge box}
\end{figure*}

Furthermore, the visualization in Fig.~\ref{fig:enlarge box} demonstrates that our refined bounding boxes better encapsulate object boundaries and contextual details, contributing to improved global and local understanding of the camouflaged targets.

\begin{table}[h!]
    \footnotesize
    \centering
    \caption{Ablation studies of different prompt strategies. Here, box denotes using minimum enclosing bounding box; rbox denotes using the refined bounding box proposed by our \ourAlgorithm.}
    \tabcolsep=0.25cm
    \begin{tabular}{c|cccccc}
    \toprule
      Model & $\mathcal{S}_\alpha\uparrow$ &$F_\beta^w\uparrow$ & $E_{\phi} \uparrow$ & $\mathcal{M}\downarrow$ & mDice$\uparrow$ & mIoU$\uparrow$ \\
     \midrule
     1p & 0.753 & 0.573 & 0.822 & 0.008 & 0.588 & 0.515\\
     3p & 0.750 & 0.572 & 0.833 & 0.008 & 0.588 & 0.512\\
     5p & 0.757 & 0.582 & 0.841 & 0.008 & 0.597 & 0.522\\
     7p & 0.740 & 0.551 & 0.815 & 0.009 & 0.567 & 0.495\\
     9p & 0.739 & 0.548 & 0.815 & 0.009 & 0.566 & 0.493\\
     \midrule
     box+5p & 0.708 & 0.481 & 0.771 & 0.008 & 0.498 & 0.430\\
     rbox+5p & \textbf{0.765} & \textbf{0.607} & \textbf{0.848} & \textbf{0.014} & \textbf{0.620} & \textbf{0.542}\\
    \bottomrule
    \end{tabular}
    \label{tab: rbbox}
\end{table}

\noindent\textbf{Application on existing VCOD methods.}
To further demonstrate the generalizability of our proposed \ourAlgorithm, we integrate it into existing VCOD methods selected from recently published non-SAM-based models with publicly available code or predictions. As shown in Table~\ref{tab: Model+AMPR}, incorporating \ourAlgorithm~into these models consistently improves their performance compared to their original versions. This highlights the effectiveness of \ourAlgorithm~in significantly enhancing the capabilities of VCOD models, even when the baseline models are not strong. 
Moreover, when applied to a stronger model, such as our proposed \ournet, \ourAlgorithm~delivers even more remarkable results, underscoring the designed components in our \ournet. 

\begin{figure}[h]
\begin{center}
\includegraphics[width=.478\textwidth]{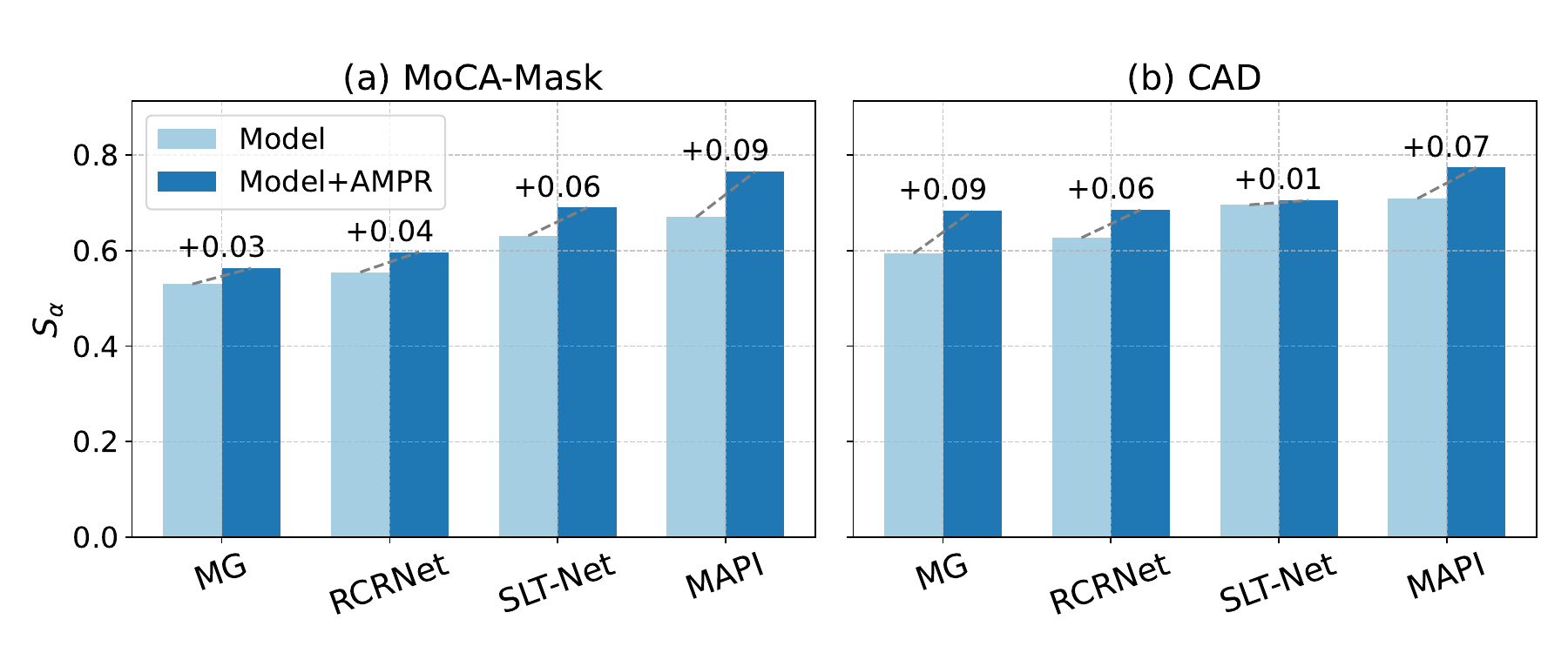}
\end{center}
\vspace{-8pt}
\caption{Illustrations of $S_{\alpha}$ change between Model and Model+AMPR.}
\label{fig:Sa-change}
\end{figure}

The localization ability of the model can be gauged by improving the quality of the mask from coarse to final refinement. Thus, we visualize the $S_{\alpha}$ improvements brought by our \ourAlgorithm~in Fig.~\ref{fig:Sa-change}. 
Compared with other methods, integrating \ourAlgorithm~into our MAPI yields more substantial gains, demonstrating that the proposed prompt inducer possesses stronger capabilities in object localization. Although MG shows slightly higher relative improvement on CAD, its final segmentation performance remains inferior to ours, indicating that many of its perceived camouflaged objects are inaccurate and lead to erroneous segmentations. In contrast, our model achieves superior performance in both camouflaged perception and localization, making its overall gains and baseline predictions highly competitive.

In summary, by leveraging different networks to generate initial mask predictions for prompt induction, the integration of \ourAlgorithm~ensures temporal consistency across predictions and achieves superior segmentation of camouflaged objects.

\begin{table*}[h!]
\centering
\caption{Quantitative comparisons of the VCOD models with their VCOD+\ourAlgorithm~ counterparts on MoCA-Mask and CAD datasets.}
\scalebox{0.93}{
\begin{tabular}{l|cccccc|cccccc}
\toprule
\multirow{2}{*}{Model} & \multicolumn{6}{c|}{MoCA-Mask~\cite{cheng2022implicit}}  & \multicolumn{6}{c}{CAD~\cite{bideau2016s}} \\ \cline{2-13} 
& $S_\alpha \uparrow$ & $F_{\beta}^{w} \uparrow$ & $E_{\phi} \uparrow$ & $\mathcal{M} \downarrow$ & mDice$\uparrow$ & mIoU$\uparrow$ & $S_\alpha \uparrow$ & $F_{\beta}^{w} \uparrow$ & $E_{\phi} \uparrow$ & $\mathcal{M} \downarrow$ & mDice$\uparrow$ & mIoU$\uparrow$ \\ 
\midrule
MG~\cite{yang2021selfsupervised} & 0.530 & 0.168 & 0.561 & 0.067 & 0.181 & 0.127 & 0.594 & 0.336 & 0.691 & 0.059 & 0.368 & 0.268 \\ 
\rowcolor{mygray}
MG+\text{\ourAlgorithm} & 0.563 & 0.177 & 0.618 & 0.039 & 0.201 & 0.138 & 0.683 & 0.469 & 0.765 & 0.030 & 0.465 & 0.402 \\ 
\midrule
RCRNet~\cite{yan2019semi} & 0.555 & 0.138 & 0.527 & 0.033 & 0.171 & 0.116 & 0.627 & 0.287 & 0.666 & 0.048 & 0.309 & 0.229 \\
\rowcolor{mygray}
RCRNet+\text{\ourAlgorithm} & 0.597 & 0.242 & 0.532 & 0.030 & 0.272 & 0.216 & 0.685 & 0.469 & 0.753 & 0.029 & 0.464 & 0.401 \\
\midrule
SLT-Net~\cite{cheng2022implicit} & 0.631 & 0.311 & 0.759 & 0.027 & 0.360 & 0.272 & 0.696 & 0.481 & 0.845 & 0.030 & 0.493 & 0.401 \\
\rowcolor{mygray}
SLT-Net+\text{\ourAlgorithm} & 0.690 & 0.454 & 0.810 & 0.021 & 0.478 & 0.401 & 0.705 & 0.516 & 0.805 & 0.035 & 0.515 & 0.415 \\
\midrule
\textbf{\ournet} (\textit{Ours}) &0.670 & 0.376 & 0.782 & 0.014 & 0.413 & 0.330 & 0.709 & 0.513 & 0.824 & 0.029 & 0.530 & 0.413 \\
\rowcolor{mygray}
\textbf{\ournet}+\text{\ourAlgorithm} & {0.765} & {0.607} & {0.848} & {0.007} & {0.620} & {0.542} & {0.774} & {0.652} & {0.852} & {0.018} & {0.647} & {0.543} \\ 
\bottomrule
\end{tabular}
}
\label{tab: Model+AMPR}
\end{table*}

\noindent\textbf{Discussion of SAM2 and SAM2-Adaptor.}
\begin{table*}[h] 
\centering
\caption{Quantitative comparison with SAM2-adapter for VCOD.}
\label{tab: Quantitative_Segmentation_Result_VCOD}
\scalebox{0.92}{
\begin{tabular}{c|cccccc|cccccc}
\toprule
\multirow{2}{*}{Model} & \multicolumn{6}{c|}{MoCA-Mask \cite{cheng2022implicit}} & \multicolumn{6}{c}{CAD \cite{bideau2016s}} \\ \cline{2-13} 
& $S_\alpha \uparrow$ & $F_{\beta}^{w} \uparrow$ & $E_{\phi} \uparrow$ & $\mathcal{M} \downarrow$ & mDice$\uparrow$ & mIoU$\uparrow$ & $S_\alpha \uparrow$ & $F_{\beta}^{w} \uparrow$ & $E_{\phi} \uparrow$ & $\mathcal{M} \downarrow$ & mDice$\uparrow$ & mIoU$\uparrow$ \\ 
\midrule
SAM2~\cite{ravi2024sam2} & 0.358 & 0.054 & 0.354 & 0.379 & 0.080 & 0.043 & 0.235 & 0.055 & 0.216 & 0.663 & 0.093 & 0.050 \\ 
SAM2-Adapter~\cite{chen2024sam2-adapter} & 0.569 & 0.162 & 0.586 & 0.041 & 0.213 & 0.144 & 0.650 & 0.387 & 0.746 & 0.043 & 0.442 & 0.329 \\ 
\textbf{\ourmodel} (\textit{Ours}) & \textbf{0.765} & \textbf{0.607} & \textbf{0.848} & \textbf{0.007} & \textbf{0.620} & \textbf{0.542} & \textbf{0.774} & \textbf{0.652} & \textbf{0.852} & \textbf{0.018} & \textbf{0.647} & \textbf{0.543} \\ 
\bottomrule
\end{tabular}
}
\end{table*}
SAM-Adapter~\cite{chen2023sam-adapter} and SAM2-Adapter~\cite{chen2024sam2-adapter} are advanced techniques that adapt the pre-trained segmentation models SAM and SAM2 to downstream tasks, thereby improving task-specific performance. We evaluate their applicability in camouflaged scenarios. As reported in Table~\ref{tab: Quantitative_Segmentation_Result_VCOD}, SAM2-Adapter attains an $S_{\alpha}$ of 0.569 on the VCOD benchmark, which is substantially lower than that of recent state-of-the-art VCOD models. In contrast, on the COD task, SAM2-Adapter not only surpasses SAM-Adapter but also outperforms several recent COD methods, as further highlighted in~\cite{chen2024sam2-adapter}, making it competitive in that setting.

This performance discrepancy between COD and VCOD can be attributed to two dataset-specific factors. First, COD provides 4,040 diverse training images, facilitating comprehensive feature learning, whereas VCOD consists of only 71 short video sequences with substantial redundancy, thereby limiting feature diversity and generalization. Second, COD offers full-HD resolution images that preserve fine-grained object boundaries, while VCOD’s lower resolutions (720p/360p) introduce boundary blurring and reduce discriminative details. Since SAM2-Adapter primarily exploits image-level features rather than motion cues, this limitation becomes more pronounced in VCOD. Therefore, relatively speaking, directly prompting SAM2 without parameter adaptation is a preferable choice for addressing VCOD.

\subsection{Module Attribution Analysis}
\label{sec:module_analysis}

To quantitatively validate our claim that localization is the primary bottleneck and to clarify the role of each component, we conduct a module-level attribution analysis with six controlled configurations (Table~\ref{tab: attribution}).

Specifically, \#1 denotes the full model (CamoSAM2). 
\#2 improves localization quality by correcting erroneous MAPI predictions, where points falling outside the target are adjusted to lie within the object region, while keeping the rest unchanged.
\#3--\#5 remove AMPR and instead use fixed multi-frame prompts (the first, middle, and last three frames, respectively), where each frame is provided with five random points and a minimum enclosing bounding box as SAM2 prompts. 
\#6 represents an oracle setting, where the initial masks are replaced with ground-truth masks for the first three frames, together with five random points and bounding box prompts, while removing both MAPI and AMPR.

\paragraph{Effect of MAPI (Localization Quality)}
Comparing \#1 and \#2, we observe consistent performance gains across datasets. For example, on CAD, the mIoU increases from 0.543 to 0.596. This demonstrates that improving localization alone leads to substantial performance improvements, indicating that localization quality plays a dominant role.

\paragraph{Effect of AMPR (Prompt Refinement)}
Comparing \#1 with \#3--\#5, we observe significant performance drops when AMPR is removed, even though the same number of prompt frames is preserved. For instance, on MoCA-Mask, the mIoU decreases from 0.542 to as low as 0.329. This confirms that the performance gain is not due to the use of multiple prompts, but rather the effectiveness of the proposed adaptive refinement strategy. Moreover, AMPR provides substantial gains only when localization is reasonably accurate, suggesting that its effectiveness is conditioned on reliable localization.

\paragraph{Effect of SAM2 Propagation (Upper Bound Analysis)}
In the oracle setting (\#6), where perfect initialization is provided using ground-truth masks, the performance remains limited (e.g., mIoU = 0.511 on MoCA-Mask). This indicates that even with accurate initialization, SAM2 propagation still struggles in challenging scenarios such as severe occlusion and small/thin objects, revealing its inherent limitations.

These results reveal a clear causal relationship: improving localization leads to the most significant performance gains, AMPR further enhances segmentation quality by refining prompts under accurate localization, while SAM2-based propagation is a secondary factor with limited impact under challenging conditions.

\begin{table}[h] 
\centering
\caption{Attribution of each module.}
\label{tab: attribution}
\tabcolsep=0.27cm
\begin{tabular}{c|ccc|ccc}
\toprule
\multirow{2}{*}{Model} & \multicolumn{3}{c|}{MoCA-Mask \cite{cheng2022implicit}} & \multicolumn{3}{c}{CAD \cite{bideau2016s}} \\ \cline{2-7} 
& $S_\alpha \uparrow$ & $F_{\beta}^{w} \uparrow$  & mIoU$\uparrow$ & $S_\alpha \uparrow$ & $F_{\beta}^{w} \uparrow$  & mIoU$\uparrow$ \\ 
\midrule
\#1 & 0.765 & 0.607 & 0.542 & 0.774 & 0.652 & 0.543 \\ 
\midrule
\#2 & 0.781 & 0.611 & 0.549 & 0.800 & 0.709 & 0.596
\\ 
\midrule
\#3 & 0.660 & 0.362 & 0.329 & 0.769 & 0.666 & 0.550 
\\ 
\#4 & 0.690 & 0.435 & 0.405 & 0.737 & 0.581 & 0.489 
\\  
\#5 & 0.667 & 0.392 & 0.362 & 0.740 & 0.584 & 0.503 
\\ 
\midrule
\#6 & 0.751 & 0.586 & 0.511 & 0.795 & 0.710 & 0.597 
\\ 
\bottomrule
\end{tabular}
\end{table}

\subsection{Cross-Task Generalization}
To evaluate cross-task generalization, we conduct experiments on several standard Video Object Segmentation (VOS) benchmarks (Table~\ref{tab:generalization}). Our CamoSAM2 achieves competitive results on DAVIS\cite{perazzi2016benchmark}, ViSal\cite{wang2015consistent}, and SegV2\cite{li2013video}, demonstrating strong adaptability to general scenarios involving complex motion.
However, we observe a performance bottleneck on the FBMS dataset. This is primarily because FBMS features sparse annotations (every 20 frames), which results in significant temporal gaps and weakened motion continuity. Since our MAPI module and the SAM2-based propagation rely heavily on dense temporal cues to distinguish objects from backgrounds, such sparsity limits their ability to accurately track and refine masks over time. Nevertheless, the consistent gains on other benchmarks underscore the robustness of our approach in standard, continuous video sequences.

\begin{table}[h]
\caption{
Comparisons of our model on VOS benchmark. The best results are \textbf{bolded} for highlighting.
}
\label{tab:generalization}
    \renewcommand\arraystretch{1.1}
	\begin{center}
		\scalebox{0.80}{
			\begin{tabular}{c|cc|cc|cc|cc}
				\toprule  
				\multirow{2}*{Method} &
				\multicolumn{2}{c|}{DAVIS\cite{perazzi2016benchmark}} & \multicolumn{2}{c|}{FBMS\cite{ochs2013segmentation}} &  \multicolumn{2}{c|}{ViSal\cite{wang2015consistent}} &
				\multicolumn{2}{c}{SegV2\cite{li2013video}}
				\\
				\cline{2-9}
	        {~}&
				\textit{$\mathcal{S}_\alpha\uparrow$}  & $\mathcal{M}\downarrow$ &
			 \textit{$\mathcal{S}_\alpha\uparrow$}  & $\mathcal{M}\downarrow$ &
				\textit{$\mathcal{S}_\alpha\uparrow$}  & $\mathcal{M}\downarrow$  & \textit{$\mathcal{S}_\alpha\uparrow$}   & 
				$\mathcal{M}\downarrow$\\
				\toprule  
				RCR\cite{yan2019semi} & 0.886  & 0.027   & 0.872    & 0.053 & -   & - & - & -\\
				CAS\cite{ji2020casnet} & 0.873  & 0.032  & 0.856 & 0.056  & -  & - & 0.820 & 0.029 \\
				DFNet & -  & 0.018 & -  & 0.054  & -  & 0.017  & - & -\\
				ReuseVOS$\dag$\cite{park2021learning}& 0.883   & 0.019  & 0.888  & \textbf{0.027}  & 0.928  & 0.020  & 0.844  & 0.025 \\
				TransVOS$\dag$\cite{mei2021transvos}& 0.885  & 0.018  & 0.867 & 0.038  & 0.917 & 0.021  & 0.816  & 0.024 \\
                UFO~\cite{su2023unified} & 0.874 & 0.032  & 0.868  & 0.041  & 0.940  & 0.012  & 0.836  & 0.057 \\
                MAMNet~\cite{zhao2024motion}& 0.897 & 0.020  & \textbf{0.894} & 0.032  & 0.947 & 0.012  & 0.886 & 0.014 \\
				\midrule  
				\textbf{CamoSAM2} & \textbf{0.899}  & \textbf{0.016}  & 0.880  & 0.035  & \textbf{0.949} & \textbf{0.012}  & \textbf{0.890} & \textbf{0.013} \\			

				\bottomrule 
		\end{tabular}}    
\end{center}
\vspace{-8pt}
\end{table}

\subsection{Robustness to Random Sampling}
Randomness in our framework arises from point prompt sampling, while pivotal-frame selection is deterministic. To assess its impact, we perform multiple inference runs with different random seeds. As reported in Table~\ref{tab:seed_robustness}, the performance variation is minimal (std $\leq$ 0.003), demonstrating that the proposed method is robust and insensitive to random sampling. This is attributed to the constrained sampling space provided by refined masks and the aggregation effect of multi-point prompting.

\begin{table}[h]
  \centering
  \caption{Robustness analysis under different random seeds. We report the mean, standard deviation (Std).}
  \label{tab:seed_robustness}
  \tabcolsep=0.45cm
  \begin{tabular}{lccc}
    \toprule
    Dataset & Mean ($S_\alpha$) & Std & Range \\
    \midrule
    MoCA-Mask & 0.765 & 0.002 & [0.760, 0.769] \\
    CAD       & 0.774 & 0.003 & [0.771, 0.779] \\
    \bottomrule
  \end{tabular}
\end{table}

\subsection{Analysis of Object-Number Estimation}
We analyze the object-number estimation in AMPR on MoCA-Mask and DAVIS (Table~\ref{tab:object-number}). It is worth noting that existing video segmentation benchmarks, including VCOD and VOS datasets, are predominantly evaluated in a per-object manner, where each sequence corresponds to a single target. Therefore, although DAVIS contains multiple objects at the video level, it is treated as a single-object setting during evaluation.
On MoCA-Mask, the predicted object number (APN = 1.7) shows a moderate over-estimation. This is beneficial, as it helps capture ambiguous regions and incomplete structures of camouflaged objects, leading to improved mIoU and $S_\alpha$ compared to enforcing a strict single-object constraint.
On DAVIS, where object appearance is clearer and localization is more accurate, the predicted number (APN = 1.1) is closer to the ground truth. In this case, the performance gain is smaller but remains consistent.
Overall, the proposed estimation behaves adaptively: it allows flexible over-estimation in challenging camouflaged scenarios while remaining accurate when object boundaries are clear, demonstrating its reliability under the current evaluation protocol.

\begin{table}[h]
\centering
\caption{Analysis of the effect of object-number estimation. We report both the Average Predicted Number (APN) of objects and the segmentation performance. Here, ``\#1'' denotes the proposed AMPR method with object-number estimation, while ``\#2'' refers to the variant without this process. }
\label{tab:object-number}
\tabcolsep=0.12cm
    \begin{tabular}{c|cccc|cccc}
        \toprule  
        \multirow{2}*{Setting} &
        \multicolumn{4}{c|}{MoCA-mask} & \multicolumn{4}{c}{DAVIS} \\
        \cline{2-9}
    {~}&
         APN & $S_\alpha \uparrow$ & $F_{\beta}^{w} \uparrow$ & mIoU$\uparrow$ & APN & $S_\alpha \uparrow$ & $F_{\beta}^{w} \uparrow$ & mIoU$\uparrow$\\
        \toprule  
		\#1 & 1.7 & 0.765 & 0.607 & 0.542 & 1.1 & 0.899 & 0.865 & 0.801\\		
		\#2 & 1.0 & 0.756 & 0.581 & 0.511 & 1.0 & 0.891 & 0.855 & 0.794\\
        \bottomrule 
\end{tabular} 
\vspace{-8pt}
\end{table}

\subsection{Failure Cases}\label{sec:failure_cases}
\begin{figure}[h]
\begin{center}
\includegraphics[width=.478\textwidth]{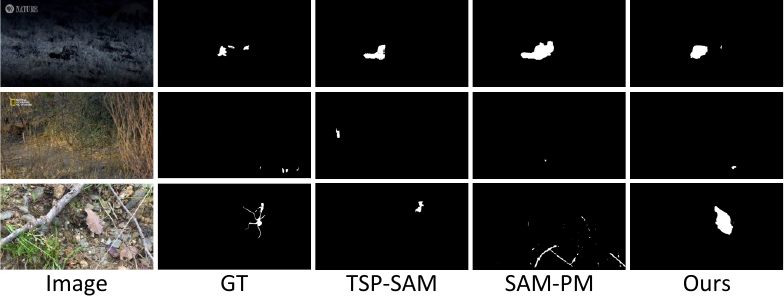}
\end{center}
\vspace{-8pt}
\caption{Failure cases of our \ourmodel~with the two most recent models.}
\label{fig:failure}
\end{figure}

Despite the strong performance of our method on standard VCOD benchmarks, several failure cases highlight limitations that merit further exploration. As illustrated in Fig.~\ref{fig:failure}, the first two rows depict scenarios involving severe occlusions—such as objects moving behind large obstacles—which lead to temporal discontinuities or complete disappearance of the target in some frames, ultimately resulting in false positives or missed detections.
The last row presents a challenging case involving a thin or small-scale object, where the model struggles to capture the fine contours due to its limited spatial resolution or attention granularity.
These observations suggest directions for future work, such as integrating additional modalities (e.g., depth or thermal cues) or designing specialized modules to better handle object disappearance and fine-scale structure under camouflage conditions.

\begin{table}[h] 
\centering
\caption{Failure analysis on different challenging cases. We report the failure rates of MoCA-Mask and CAD  under two typical challenging scenarios.}
\label{tab: Failure_type}
\tabcolsep=0.25cm
\begin{tabular}{c|cc}
\toprule
Types & MoCA-Mask \cite{cheng2022implicit} & CAD \cite{bideau2016s} \\ 
\midrule
severe occlusion & 1.5\% & 5.5\% \\ 
small/thin objects & 7.3\% & 1.9\% \\ 
\bottomrule
\end{tabular}
\end{table}

\paragraph{Failure Type Statistics}
To complement the qualitative analysis, we further provide a compact quantitative summary of failure types on MoCA-Mask and CAD (Table~\ref{tab: Failure_type}). We categorize failure cases into two representative types: (1) severe occlusion, and (2) small or thin objects.
As shown in Table~\ref{tab: Failure_type}, failures caused by small or thin objects are more frequent on MoCA-Mask (7.3\%), while severe occlusion is more prominent on CAD (5.5\%). This difference reflects the distinct challenges of the two datasets. In particular, small object structures are more vulnerable to prompt inaccuracies, whereas occlusion primarily affect temporal consistency.
These observations are consistent with our qualitative analysis and further support the necessity of accurate localization and robust propagation under challenging conditions.

\section{Conclusion}
In this paper, we introduce \ourmodel, a novel framework that employs a motion-appearance induced, auto-refining prompt approach to achieve reliable and precise video camouflaged object detection. Our method begins with a motion-appearance prompt inducer module, enabling the detection of camouflaged objects without user-provided prompts.
Moreover, we propose an innovative video-based adaptive multi-prompts refinement strategy, which autonomously produces reliable and robust prompts for high-precision segmentation results, without increasing the number of training parameters. Notably, this strategy is composed of three meticulously designed steps, making it highly suitable for real-world applications.
Experimental results on benchmark datasets show that our \ourmodel~significantly outperforms existing state-of-the-art methods on evaluation metrics and inference speed. 
Our proposed method provides a novel perspective for adapting SAM2 to the VCOD task. Looking ahead, future work can focus on enhancing SAM2 by endowing it with intrinsic camouflage perception capabilities, enabling it to directly perform fully automated, interaction-free VCOD.

\bibliographystyle{IEEEtran}
\bibliography{main}

\end{document}